\pdfoutput=1

\documentclass[11pt]{article}

\usepackage{acl}
\usepackage{times}
\usepackage{latexsym}
\usepackage{amsthm,amsmath,amssymb}
\usepackage{mathrsfs}
\usepackage{float}
\usepackage{CJKutf8}

\usepackage{multirow} 
\usepackage{graphicx}
\usepackage{subfigure}
\usepackage{stfloats}
\graphicspath{ {./pic} }
\usepackage{multicol}
\usepackage{booktabs}
\usepackage[ruled,linesnumbered]{algorithm2e}
\SetKwRepeat{Do}{do}{while}%
\SetKwInOut{Input}{Output}
\usepackage{float}
\usepackage{bm}
\usepackage{enumitem}
\usepackage[T1]{fontenc}

\usepackage[utf8]{inputenc}

\usepackage{microtype}

\title{Generating Authentic Adversarial Examples beyond Meaning-preserving with Doubly Round-trip Translation}

\author{Siyu Lai\textsuperscript{1}\thanks{ \ \ Work was done when Siyu were interning at Pattern Recognition Center, WeChat AI, Tencent Inc, China.},
Zhen Yang\textsuperscript{2} , 
Fandong Meng\textsuperscript{2},
Xue Zhang \textsuperscript{1},
\textbf{Yufeng Chen}\textsuperscript{1}\thanks{ \ \ Yufeng Chen is the corresponding author.}, \\ 
\textbf{Jinan Xu}\textsuperscript{1} and \textbf{Jie Zhou}\textsuperscript{2}\\
\textsuperscript{1}Beijing Key Lab of Traffic Data Analysis and Mining, \\
Beijing Jiaotong University, Beijing, China \\
\textsuperscript{2}Pattern Recognition Center, WeChat AI, Tencent Inc, China \\
\texttt{\{siyulai,xue\_zhang,chenyf,jaxu\}@bjtu.edu.cn}, \\
\texttt{\{zieenyang,fandongmeng,withtomzhou\}@tencent.com} \\}

\begin{document}
\maketitle
\begin{abstract}
Generating adversarial examples for Neural Machine Translation (NMT) with single Round-Trip Translation (RTT) has achieved promising results by releasing the meaning-preserving restriction. However, a potential pitfall for this approach is that we cannot decide whether the generated examples are adversarial to the target NMT model or the auxiliary backward one, as the reconstruction error through the RTT can be related to either. To remedy this problem, we propose a new criterion for NMT adversarial examples based on the Doubly Round-Trip Translation (DRTT). Specifically, apart from the source-target-source RTT, we also consider the target-source-target one, which is utilized to pick out the authentic adversarial examples for the target NMT model. Additionally, to enhance the robustness of the NMT model, we introduce the masked language models to construct bilingual adversarial pairs based on DRTT, which are used to train the NMT model directly. Extensive experiments on both the clean and noisy test sets (including the artificial and natural noise) show that our approach substantially improves the robustness of NMT models.
\end{abstract}

\section{Introduction}
In recent years, neural machine translation (NMT) \cite{cho2014learning,bahdanau2014neural,vaswani2017attention} 
has achieved rapid advancement in the translation performance \cite{yang2020csp,lu-etal-2021-attention}. However, the NMT model is not always stable enough, as its performance can drop significantly when small perturbations are added into the input sentences \cite{belinkov2017synthetic,cheng-etal-2020-advaug}. Such perturbed inputs are often referred to as adversarial examples in the literature, and how to effectively generate and utilize adversarial examples for NMT is still an open question.

\begin{figure}
\centering
\includegraphics[width=0.5\textwidth]{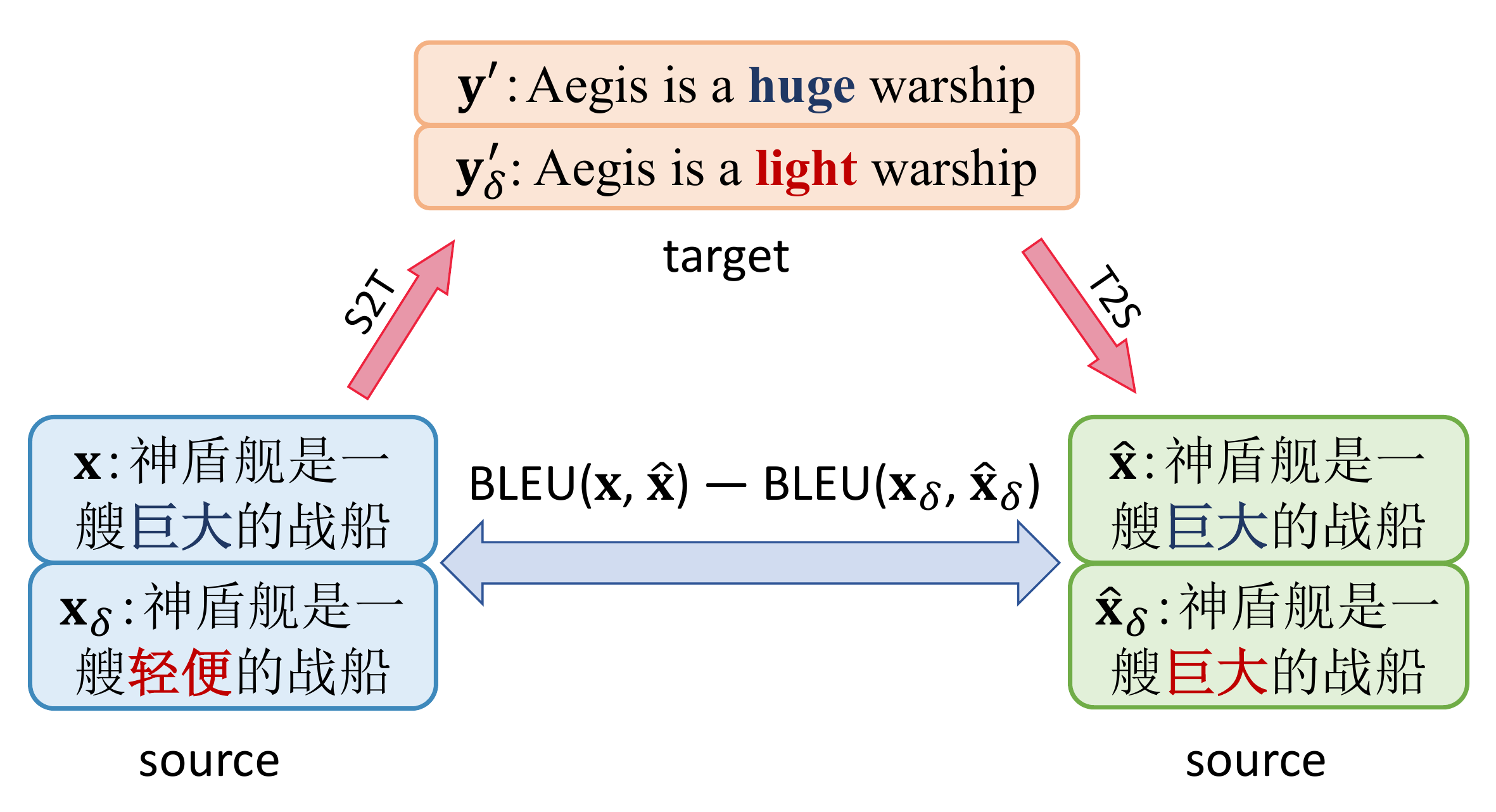}
\caption{\label{fig:abs} An example of the source-target-source RTT process on a perturbed input $\mathbf x_\delta$ by replacing ``\begin{CJK*}{UTF8}{gbsn}{巨大}\end{CJK*} (huge)'' to ``\begin{CJK*}{UTF8}{gbsn}{轻便}\end{CJK*} (light)''. } 
\end{figure}

Conventional approaches \cite{ebrahimi2018hotflip,cheng2019robust} for generating NMT adversarial examples always follow the meaning-preserving assumption, i.e., an NMT adversarial example should preserve the meaning of the source sentence but destroy the translation performance drastically \cite{michel2019evaluation,niu-etal-2020-evaluating}.  With the meaning-preserving restriction, the researchers try to add perturbations on the source inputs as small as possible to ensure the meaning of the source sentence is unchanged, which severely limits the search space of the adversarial examples. Additionally, it is much problematic to craft a minor perturbation on discrete text data, since some random transformations (e.g., swap, deletion and replacement) may change, or even reverse semantics of the text data, breaking the aforementioned meaning-preserving assumption. 
To break this limitation, \citet{zhang-etal-2021-crafting} introduce a new criterion for NMT adversarial examples: \emph{an effective NMT adversarial example imposes minor shifting on the source and degrades the translation dramatically, would naturally lead to a semantic-destroyed round-trip translation result}. Take the case in Figure \ref{fig:abs} as an example: $\mathbf x_{\mathbf \delta}$ reverses the semantics of input $\mathbf x$ by replacing ``\begin{CJK*}{UTF8}{gbsn}{巨大}\end{CJK*} (huge)'' to ``\begin{CJK*}{UTF8}{gbsn}{轻便}\end{CJK*} (light)''. Since the semantics of $\mathbf x$ and $\mathbf x_{\mathbf \delta}$ are completely different, it is unreasonable to use the original target sentence of $\mathbf x$ to evaluate the attacks directly.
Therefore, \citet{zhang-etal-2021-crafting} propose to evaluate the BLEU score between $\mathbf {x_\delta}$ and  its reconstructed sentence $\mathbf {\hat{x}_\delta}$ from the source-target-source round-trip translation (RTT), as well as the BLEU score between the original sentence $\mathbf x$ and its reconstructed sentence $ \mathbf {\hat{x}}$. 
They take the decrease between the two BLEU scores mentioned above as the adversarial effect.
Specifically, if the BLEU decrease exceeds a predefined threshold, $\mathbf{ x_\delta}$ is concluded to be an adversarial example for the target NMT model.

While achieving promising results by breaking the meaning-preserving constraint, there are two potential pitfalls in the work of \citet{zhang-etal-2021-crafting}: (1) Since the source-target-source RTT involves two stages, i.e., the source-to-target translation (S2T) performed by the target NMT model and target-to-source translation (T2S) performed by an auxiliary backward NMT model, we cannot decide whether the BLEU decrease is really caused by the target NMT model. As we can see from the example in Figure \ref{fig:abs}, the translation from $\mathbf {x_\delta}$ to $\mathbf {y'_\delta}$ is pretty good, but the translation from $\mathbf{ y'_\delta}$ to $\mathbf {\hat {x}_\delta}$ is really poor. We can conclude that the BLEU decrease is actually caused by the auxiliary backward model and thus $\mathbf {x_\delta}$ is not the adversarial example for the target NMT model. Even if  \citet{zhang-etal-2021-crafting} try to mitigate this problem by fine-tuning the auxiliary backward model on the test sets, we find this problem still remains. (2) They only generate the monolingual adversarial examples on the source side to attack the NMT model, without proposing methods on how to defend these adversaries and improve the robustness of the NMT model.

To address the issues mentioned above, we first propose a new criterion for NMT adversarial examples based on Doubly Round-Trip Translation (DRTT), which can ensure the examples that meet our criterion are the authentic adversarial examples for the target NMT model. 
Specifically, apart from the source-target-source RTT \cite{zhang-etal-2021-crafting}, we additionally consider a target-source-target RTT on the target side.
The main intuition is that an effective adversarial example for the target NMT model shall cause a large BLEU decrease on the source-target-source RTT while maintaining a small BLEU decrease on  target-source-target RTT. Based on this criterion, we craft the candidate adversarial examples with the source-target-source RTT as \citet{zhang-etal-2021-crafting}, and then pick out the authentic adversaries with the target-source-target RTT. Furthermore, to  solve the second problem, we introduce the masked language models (MLMs) to construct the bilingual adversarial pairs by performing phrasal replacement on the generated monolingual adversarial examples and the original target sentences synchronously, which are then utilized to train the NMT model directly. Experiments on both clean and noisy test sets (including five types of artificial and nature noise) show that the proposed approach not only generates effective adversarial examples, but also improves the robustness of the NMT model over all kinds of noises. To conclude, our main contributions are summarized as follows:
\begin{itemize}[leftmargin=*]
    \item{We propose a new criterion for NMT adversarial examples based on the doubly round-trip translation, which can pick out the authentic adversarial examples for the target NMT model. }
    \item{We introduce the masked language models to construct the bilingual adversarial pairs,
    which are then utilized to improve the robustness of the NMT model.}
    \item{Extensive experiments show that the proposed approach not only improves the robustness of the NMT model on both artificial and natural noise, but also performs well on the clean test sets\footnote{The code is publicly available at: \url{https://github.com/lisasiyu/DRTT}}. }
\end{itemize}

\section{Related Work}
\subsection{Adversarial Examples for NMT}
The previous approaches for constructing NMT adversarial examples  can be divided into two branches: white-box and black-box. 
The white-box approaches are based on the assumption that the architecture and parameters of the NMT model are accessible \cite{ebrahimi2018hotflip,cheng2019robust,chen2021manifold}. These methods usually achieve superior performance since they can construct and defend the adversaries tailored for the model. However, in the real application scenario, it is always impossible for us to access the inner architecture of the model. On the contrary, the black-box approaches 
never access to inner architecture and parameters of the model. In this line, \citet{belinkov2017synthetic} rely on synthetic and naturally occurring language error to generate adversarial examples and \citet{michel2019evaluation} propose a meaning-preserving method by swapping the word internal character.
Recently, \citet{zhang-etal-2021-crafting} craft adversarial examples beyond the meaning-preserving restriction with the round-trip translation and our work builds on top of it. 

\subsection{Masked Language Model}
Masked Language Model (MLM) \cite{bert,conneau2019cross} has achieved state-of-the-art results on many monolingual and cross-lingual language understanding tasks. MLM randomly masks some of the tokens in the input, and then predicts those masked
tokens. Recently, some work adopt MLM to do word replacement as a data augmentation strategy. \citet{jiao2019tinybert} leverage an encoder-based MLM to predict word replacements for single-piece words. \citet{liu2021counterfactual} construct augmented sentence pairs by sampling new source phrases and corresponding target phrases with transformer-based MLMs. Following \citet{liu2021counterfactual}, we introduce the transformer-based MLMs to construct the bilingual adversarial pairs.  The main difference between our work and \citet{liu2021counterfactual} is that we choose to mask the adversarial phrases or words at each step and \citet{liu2021counterfactual} mask the words randomly.

\begin{figure*}
\centering
\includegraphics[width=0.9\textwidth]{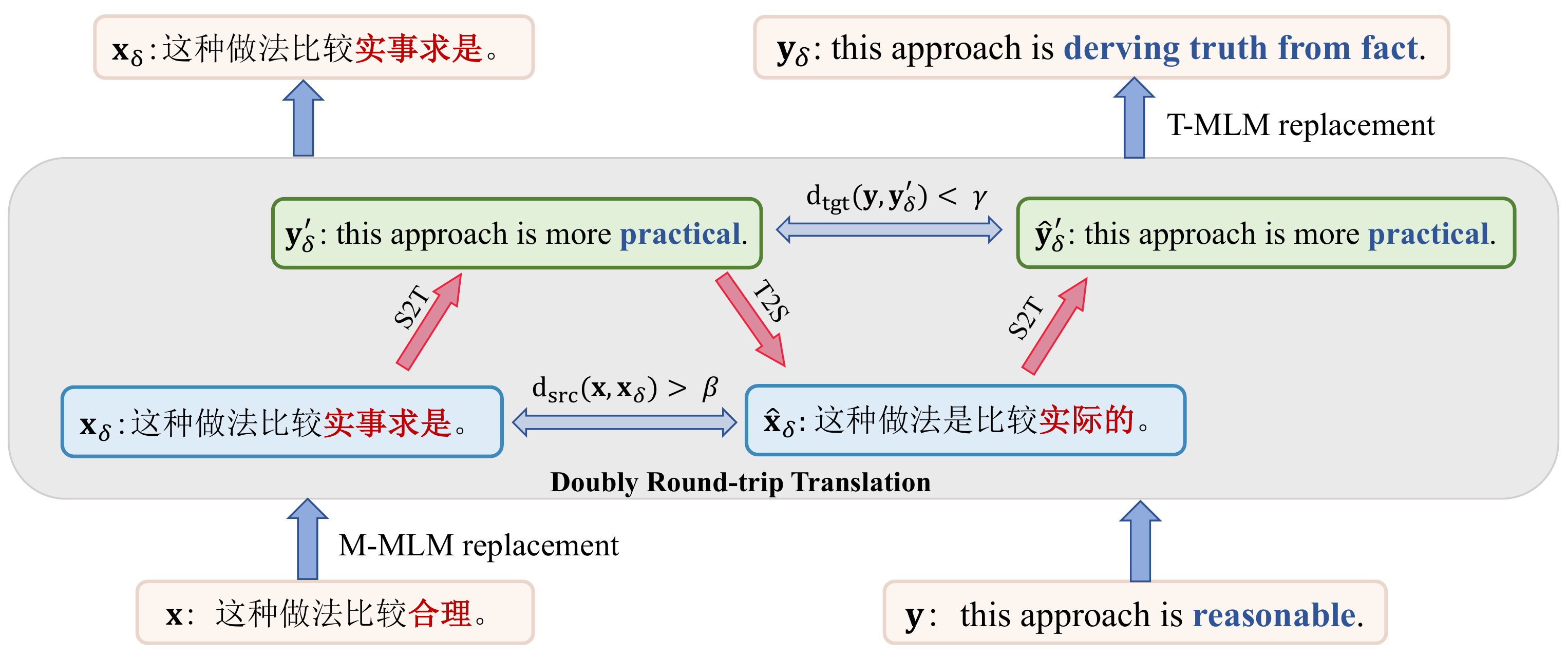}
\caption{\label{fig:1} The overview of the bilingual adversarial pair generation under the criterion of DRTT. $(\mathbf x,\mathbf y)$ denote the source and target sentence. $(\mathbf {x_\delta},
\mathbf {y_\delta})$ denote the generated bilingual adversarial pair. } 
\end{figure*}

\section{Method}
In this section, we first describe our proposed criterion for NMT adversarial examples, and then present the way of constructing the bilingual adversarial pairs.

\subsection{Adversarial Examples for NMT}
For clarity, we first introduce the traditional criteria for NMT adversarial examples, i.e., the criteria based on the meaning-preserving \citep{michel2019evaluation,karpukhin2019training} and RTT \citep{zhang-etal-2021-crafting}, and then elaborate our new criterion based on DRTT. 
We will use the following notations: $\mathbf x$ and $\mathbf y$ denotes the source and target sentence, respectively. $\mathbf {x_\delta}$ and $\mathbf {y_\delta}$ denote the perturbed version of $\mathbf x$ and $\mathbf y$, respectively. $f(\cdot)$ is the forward translation process performed by the target NMT model and $g(\cdot)$ is the backward translation process performed by the auxiliary backward NMT model. ${\rm sim}(\cdot,\cdot)$ is a function for evaluating the similarity of two sentences, and we use BLEU \citep{papineni2002bleu} as the similarity function.

\paragraph{Criterion based on meaning-preserving.}
Suppose $\mathbf {y'}=f(\mathbf x)$ and $\mathbf {y_\delta'}=f(\mathbf { x_\delta})$ is the forward translation of the input $\mathbf x$ and its perturbed version $\mathbf{x_\delta}$, respectively. 
$\mathbf x_\delta$ is an adversarial examples when it meets:
\begin{equation}
\left\{
    \begin{array}{lr}
    {\rm sim}(\mathbf x,\mathbf{x_\delta}) > \eta, & \\
{\rm sim}(\mathbf y,\mathbf{y'})-{\rm sim}(\mathbf y,\mathbf{y'_{\delta}}) > \alpha,
    \end{array}
\right.
\end{equation}
where $\eta$ is a threshold to ensure a high similarity between $\mathbf {x_\delta}$ and $\mathbf x$, so that they can meet the meaning-preserving restriction. A larger $\alpha$ indicates a more strict criterion of the NMT adversarial example.
 
\paragraph{Criterion based on RTT.}
\citet{zhang-etal-2021-crafting} point out that the perturbation $\delta$ may change, even reverse the meaning of $\mathbf x$,
so it is incorrect to use $\mathbf{y}$ as a target sentence to measure the semantic alteration on the target side. Therefore, they introduce the criterion based on RTT which gets rid of the meaning-preserving restriction. The percentage decrease of similarity between $\mathbf{x}$ and $\mathbf{x_\delta}$ through the source-target-source RTT is regarded as the adversarial effect ${\rm d_{src}}(\mathbf{x},\mathbf{x_\delta})$, is calculated as:
\begin{equation}
   {\rm d_{src}}(\mathbf{x},\mathbf{x_\delta})=\frac{{}{\rm sim}(\mathbf x,\mathbf{\hat{x}})-{\rm sim}(\mathbf{x_{\delta}},\mathbf{\hat{x}_\delta})}{{\rm sim}(\mathbf x,\mathbf{\hat{x})}},\label{func:2}
\end{equation}
where $\mathbf{\hat{x}}$ and $\mathbf{{\hat{x}}_\delta}$ are reconstructed sentences generated with source-target-source RTT: $\mathbf{\hat{x}}=g(f(\mathbf x))$, ${\mathbf{\hat{x}_\delta}}=g(f(\mathbf{x_\delta}))$. 
A large ${\rm d_{src}}(\mathbf{x},\mathbf{x_\delta})$ indicates that the perturbed sentence $\mathbf{x_\delta}$ can not be well reconstructed by RTT when compared to the reconstruction quality of the original source sentence $\mathbf x$, so $\mathbf{x_\delta}$ is likely to be an adversarial example.

\paragraph{ Criterion based on DRTT.} 
\label{sec:2.3}
In Eq.(\ref{func:2}), $\rm sim(\mathbf x,\mathbf{\hat{x}})$ is a constant value given the input $\mathbf x$ and the NMT models. Therefore, the ${\rm d_{src}}(\mathbf{x},\mathbf{x_\delta})$ is actually determined by $-\rm sim(\mathbf{x_{\delta}},\mathbf{{\hat{x}_\delta}})$, which can be interpreted as the reconstruction error between $\mathbf{x_{\delta}}$ and $\mathbf{{\hat{x}_\delta}}$. As we mentioned above, the reconstruction error can be caused by two independent translation processes: the forward translation process $f(\cdot)$ 
performed by the target NMT model
and the backward translation process $g(\cdot)$
performed by the auxiliary backward model. Consequently, there may be three occasions when we get a large ${\rm d_{src}}(\mathbf{x},\mathbf{x_\delta})$: 1) A large semantic alteration in $f(\mathbf{x_{\delta}})$ and a small semantic alteration in $g(\mathbf{y'_{\delta}})$; 2) A large semantic alteration in $f(\mathbf{x_{\delta}})$ and a large alteration in $g(\mathbf{y'_{\delta}})$; 3) A small semantic alteration in $f(\mathbf{x_{\delta}})$ and a large alteration in $g(\mathbf{y'_{\delta}})$. We can conclude $\mathbf{x_{\delta}} $ is an adversarial example for the target NMT model in occasion 1 and 2, but not in occasion 3. Therefore, the criterion based on RTT may contain many fake adversarial examples. 

To address this problem, we add a target-source-target RTT starting from the target side. The percentage decrease of the similarity between $\mathbf{y}$ and $\mathbf{y'_\delta}$ through the target-source-target RTT, denoted as ${\rm {d_{tgt}}}(\mathbf{y},\mathbf{y'_\delta})$, is calculated as:
\begin{equation}
    {\rm {d_{tgt}}}(\mathbf{y},\mathbf{y'_\delta})= \frac{{\rm sim}(\mathbf y,\mathbf{\hat{y}})-{\rm sim}(\mathbf{y'_{\delta}},\mathbf{\hat{y}'_{\delta}})}{{\rm sim}(\mathbf y,\mathbf{\hat{y}})}, \label{func:3}
\end{equation}
where $\mathbf{\hat{y}}=f(g(\mathbf y))$ and $\mathbf{\hat y'_{\delta}} = f(g(\mathbf{y'_\delta}))$ are reconstructed sentences generated with the target-source-target RTT.
We take both ${\rm d_{src}}(\mathbf{x},\mathbf{x_\delta})$ and ${\rm {d_{tgt}}}(\mathbf{y},\mathbf{y'_\delta})$ into consideration and define $\mathbf x_\delta$ as an adversarial examples when it meets:
\begin{equation}
\left\{
    \begin{array}{lr}
    {\rm {d_{src}}}(\mathbf{x},\mathbf{x_\delta}) > \beta, & \\
    {\rm {d_{tgt}}}(\mathbf{y},\mathbf{y'_\delta}) < \gamma, \label{func:4}
    \end{array}
\right.
\end{equation}
where $\beta$ and $\gamma$ are thresholds ranging in $[-\infty,1]$ \footnote{It is possible that the reconstruction quality of the perturbed sentence is higher than the original one.}. The interpretation of this criterion is intuitive: if ${\rm {d_{tgt}}}(\mathbf{y},\mathbf{y'_\delta})$ is lower than $\gamma$, we can conclude that the reconstruction error between $\mathbf{y'_{\delta}}$ and $\mathbf{\hat y'_{\delta}}$ is very low. Namely, we can ensure a small semantic alteration of $g(\mathbf{y'_{\delta}})$. Therefore, if ${\rm {d_{src}}}(\mathbf{x},\mathbf{x_\delta})$ is larger than $\beta$, we can conclude the BLEU decrease through the source-target-source RTT is caused by the target NMT model, so that we can conclude $\mathbf x_\delta$ is an authentic adversarial example. 

\subsection{Bilingual Adversarial Pair Generation}
Since the proposed criterion breaks the meaning-preserving restriction, the adversarial examples may be semantically distant from the original source sentence. Thus, we cannot directly pair the adversarial examples with the original target sentences. 
In this section, we propose our approach for generating bilingual adversarial pairs, which performs the following three steps: 1) Training Masked Language Models: using monolingual and parallel data to train masked language models; 2) Phrasal Alignment: obtaining alignment between the source and target phrases; 3) Phrasal Replacement: generating bilingual adversarial pairs by performing phrasal replacement on the source and target sentences synchronously with the trained masked language models. The whole procedure is illustrated in Figure \ref{fig:1}.

\paragraph{Training Masked Language Models.}
We train two kinds of masked language models, namely monolingual masked language model (M-MLM) \citep{bert} and translation masked language model (T-MLM) \citep{conneau2019cross}, for phrasal replacement on the source and target sentence, respectively. The M-MLM introduces a special [MASK] token which randomly masks some of the tokens from the input in a certain probability, and predict the original masked words. Following \citet{liu2021counterfactual}, we train the M-MLM on monolingual datasets and use an encoder-decoder Transformer model \citep{vaswani2017attention} to tackle the undetermined number of tokens during generation. The T-MLM takes the identical model structure and similar training process as the M-MLM. The main difference is T-MLM relies on the parallel corpus. T-MLM concatenates parallel sentences by a special token [SEP] and only masks words on the target side. The objective is to predict the original masked words on the target side.

\paragraph{Phrasal Alignment.}
Phrasal alignment projects each phrase in the source sentence $\mathbf x$ to its alignment phrase in the target sentence $\mathbf y$. We first generate the alignment between $\mathbf x$ and $\mathbf y$ using FastAlign \citep{dyer2013simple}. Then we extract the phrase-to-phrase alignment by the phrase extraction algorithm of NLTK\footnote{\url{https://github.com/nltk/nltk/blob/develop/nltk/translate/phrase_based.py}}, and get a mapping function $p$.

\paragraph{Phrasal Replacement.}
Given the source sentence ${\mathbf x}=\{s_1,s_2,\dots,s_n\}$ and the target sentence ${\mathbf y}=\{t_1,t_2,\dots,t_m\}$, $s_i$ is the $i$-th phrase in $\mathbf x$, $t_{p(i)}$ is the $p(i)$-th phrase in $\mathbf y$ which is aligned to $s_i$ by the mapping function $p$. We construct the candidate bilingual adversarial pairs $(\mathbf x_\delta,\mathbf y_\delta)$ by performing the phrasal replacement on $(\mathbf x, \mathbf y)$ repeatedly until $c$ percentage phrases in $\mathbf x$ have been replaced. For each step, we select the phrase that yields the most significant reconstruction quality degradation.

Here, we take the replacing process for $s_i$ and $t_{p(i)}$ as an example. Considering the not attacked yet phrase $s_i$ in $\mathbf x$, we first build a candidate set $\mathcal{R}_i=\{r_i^1,r_i^2,\dots,r_i^k\}$ for $s_i$ with the prepared M-MLM. Specifically, we extract the $k$ candidate phrases with top $k$ highest  predicted probabilities by feeding $\mathbf x^{\backslash i}$ into M-MLM, where $\mathbf x^{\backslash i}$ is the masked version of $\mathbf x$ by masking $s_i$. We select the best candidate $r_i^*$ for $s_i$ as:
\begin{equation}
    r_i^* = \mathop{\arg\max}\limits_{j \in \{1,\cdots,k\}} {\rm d_{src}}(\mathbf x,\mathbf x^{\backslash i:j}),
\end{equation}
where $\mathbf x^{\backslash i:j}$ is the noised version by replacing $s_i$ with $r_i^j$.
With $s_i$ being replaced, we need to replace $t_{p(i)}$ to ensure they are still semantically aligned. 
To this end, we feed the concatenation of $\mathbf x^{\backslash i:*}$ and $\mathbf y^{\backslash p(i)}$ into T-MLM, and choose the output phrase with the highest predicted probability as the substitute phrase for $t_{p(i)}$.

Finally, to decide whether $(\mathbf x_\delta,\mathbf y_\delta)$ is an authentic bilingual adversarial pair for the target NMT model, we perform a target-source-target RTT starting from the target side and calculate ${\rm d_{tgt}}(\mathbf y,\mathbf {y'_\delta})$ between $\mathbf y'_\delta$ and its reconstruction sentence $\mathbf{\hat{y}'_\delta}$ according to Eq.(\ref{func:4}). We take $(\mathbf x_\delta,\mathbf y_\delta)$ as an authentic bilingual adversarial pair 
if ${\rm d_{src}}(\mathbf x,\mathbf{x_\delta})$ is greater than $\beta$ and ${\rm d_{tgt}}(\mathbf y,\mathbf{y'_\delta})$ is less than $\gamma$. We formalize these steps in Algorithm \ref{algorithm:alg1} in Appendix \ref{sec:appendixa}.

After generating adversarial data through the above steps, we combine it with original training data and use them to train the NMT model directly.

\section{Experimental Settings}
We evaluate our model under artificial noise in Zh$\rightarrow$En and En$\rightarrow$De translation tasks, and under natural noise in En$\rightarrow$Fr translation task.
The details of the experiments are elaborated in this section.
\subsection{Dataset}
For the Zh$\rightarrow$En task, we use the LDC corpus with 1.25M sentence pairs for training\footnote{It is extracted from LDC data, including LDC 2002E18, 2003E07, 2003E14, 2004T08 and 2005T06.}, NIST06 for validation, and NIST 02, 03, 04, 05, 08 for testing. For the En$\rightarrow$De task, we use the publicly available dataset WMT'17 En-De (5.85M) for training, and take the \emph{newstest16} and \emph{newstest17} for validation and testing, respectively. In En$\rightarrow$Fr task, we follow \citet{liu2021counterfactual} to combine the WMT'19 En$\rightarrow$Fr (36k) robustness dataset with Europarl-v7 (2M) En-Fr pairs for training. We take the development set of the MTNT \cite{michel2018mtnt} for validation and the released test set of the WMT'19 robustness task for testing. As for MLMs, we use the Chinese sentences of the parallel corpus to train the Chinese M-MLM, and use the whole parallel corpus to train Zh-En T-MLM. We train the English M-MLM with News Commentary and News Crawl 2010 (7.26M in total) monolingual corpus following \citet{liu2021counterfactual}. T-MLM for En-De and En-Fr are trained with their original parallel corpus.

\begin{table*}
\centering
\renewcommand\arraystretch{0.98}
\scalebox{0.95}{

\resizebox{\linewidth}{!}{
  \begin{tabular}{c|c|c c c | c|c c c| c}
        \toprule[1.2pt]
        \multirow{2}{*}{\textbf{Noise}} &\multirow{2}{*}{\textbf{Model}} &  \multicolumn{4}{c|}{\textbf{Zh$\rightarrow$En}} & \multicolumn{4}{c}{\textbf{En$\rightarrow$De}} \\
        \cmidrule(r){3-6}\cmidrule(r){7-10}
        \multicolumn{1}{c|}{}&\multicolumn{1}{c|}{} &  0.1 & 0.2 & 0.3 & \textbf{AVG} & 0.1 & 0.2 & 0.3 & \textbf{AVG}   \\
        \hline
        \multirow{5}{*}{\textbf{Deletion}} & \multicolumn{1}{l|}{baseline}  & 32.98 & 26.59 & 20.54 &26.70 & 19.82 & 13.71 & 9.33 & 14.29   \\
        & \multicolumn{1}{l|}{+CharSwap} & 32.94 & 26.92 & 20.46 & 26.77 & \textbf{19.92} & 13.64 & 9.30 & 14.29\\
        & \multicolumn{1}{l|}{+TCWR}  & 34.47 & 27.76 & 21.38 & 27.87 & 19.61 &	13.77 & 9.08 & 14.15  \\ 
        & \multicolumn{1}{l|}{+RTT}  & 33.84 & 27.43 & 20.74 & 27.33 & 19.61 & 13.48 & 9.27 & 14.12 \\ 
        & \multicolumn{1}{l|}{+DRTT(ours)}  & ~~~~\textbf{35.10$**$} & ~~\textbf{28.12$*$} & ~~~~\textbf{22.07$**$} & \textbf{28.43}  & 19.83 & \textbf{14.22} & \textbf{9.48} & \textbf{14.51} \\ 
        \hline
        \multirow{5}{*}{\textbf{Swap}} & \multicolumn{1}{l|}{baseline}  & 36.14 & 32.88 & 30.21 & 33.08 &  21.47 & 16.97 & \textbf{13.21} & 17.22 \\
        & \multicolumn{1}{l|}{+CharSwap} & 36.71 & 33.38 & 30.58 & 33.55 & 20.49 & 16.31 & 11.93 & 16.24\\
        & \multicolumn{1}{l|}{+TCWR}  & 37.67 & 34.15 & 31.47 & 34.43  & 20.52 & 16.31 & 12.80 & 16.54  \\ 
        & \multicolumn{1}{l|}{+RTT}  & 37.14 & 34.34 & 31.42 & 34.30 & 20.23 & 15.47 & 11.52 & 15.74 \\ 
        & \multicolumn{1}{l|}{+DRTT(ours)} & ~~\textbf{37.90$*$} & \textbf{34.65} & ~~\textbf{31.92$*$} & \textbf{34.82} & ~~~~\textbf{21.51}$**$ & ~~~~\textbf{17.36}$**$ & ~~~~12.91$**$ & \textbf{17.26}\\ 
        \hline
        \multirow{5}{*}{\textbf{Insertion}} & \multicolumn{1}{l|}{baseline} & 39.96 & 39.10 & 38.41 & 39.16 & 26.86 & \textbf{26.54} & 25.48 & 25.96  \\
        & \multicolumn{1}{l|}{+CharSwap} & 40.26 & 39.66 & 39.03 &39.65 & 26.69 & 25.79 & 25.23 & 25.90\\
        & \multicolumn{1}{l|}{+TCWR}  & 41.32 & 40.07 & 39.60 & 40.33 & 26.27 & 25.55 & 24.33 & 25.38 \\ 
        & \multicolumn{1}{l|}{+RTT} & 41.75 & 40.82 & 39.90 & 40.82& 26.18 & 25.06 & 23.68 & 24.97  \\ 
        & \multicolumn{1}{l|}{+DRTT(ours)} & \textbf{41.98} & \textbf{40.90} & ~~\textbf{40.34$*$} & \textbf{41.07} & ~~~~\textbf{27.32}$**$ & ~~~~26.40$**$ & ~~~~\textbf{25.71}$**$ & \textbf{26.48}\\ 
        \hline
        \multirow{5}{*}{\textbf{Rep src}} & \multicolumn{1}{l|}{baseline}  & 35.25 & 29.69 & 24.64 &  29.86& \textbf{21.65} & 17.40 & 14.45 & 17.83  \\
        & \multicolumn{1}{l|}{+CharSwap} & 35.01 & 30.25 & 25.27 & 30.18 & 21.56 & 17.67 & 14.60 & 17.94\\
        & \multicolumn{1}{l|}{+TCWR}  & 35.73 & \textbf{30.48} & 25.65 & \textbf{30.62} & 21.57 & \textbf{17.71} & \textbf{14.95} & \textbf{18.08}  \\ 
        & \multicolumn{1}{l|}{+RTT} & 35.63 & 30.17 & \textbf{25.86} & 30.55 & 21.06 & 17.01 & 14.36& 17.48 \\ 
        & \multicolumn{1}{l|}{+DRTT(ours)} & \textbf{35.81} & 30.18 & 25.70 &  30.56& ~~21.51$*$ & 17.22 & 14.33 & 17.69  \\ 
                \hline
        \multirow{5}{*}{\textbf{Rep both}} & \multicolumn{1}{l|}{baseline}  & 22.33 & 18.77 & 15.98 & 19.03 & 25.52 & 22.68 & 20.07 & 22.76\\
        & \multicolumn{1}{l|}{+CharSwap} & 21.99 & 18.08 & 15.77 & 18.61 & 25.18 & 22.39 & 19.98 & 22.52 \\
        & \multicolumn{1}{l|}{+TCWR} & 22.98 & 19.69 & 17.14 & 19.94 & 25.44 & 22.64 & 20.43 & 22.84\\ 
        & \multicolumn{1}{l|}{+RTT}  & 22.92 & 19.56 & 16.76 & 19.75 & 25.30 & 22.76 & 20.66 & 22.91\\ 
        & \multicolumn{1}{l|}{+DRTT(ours)}  & ~~~~\textbf{23.37$**$} & ~~~~\textbf{20.23$**$} & ~~~~\textbf{17.37$**$} &  \textbf{20.32} & ~~\textbf{26.19}$*$ & ~~~~\textbf{23.31}$**$ & \textbf{20.98} & \textbf{23.49}\\ 
    \bottomrule
    \end{tabular}
}      
}
\caption{\label{tab:2} The BLEU scores (\%) for forward-translation on noisy test sets with noise ratio 0.1, 0.2 and 0.3, and `AVG' denotes the average BLEU (\%) on all noise ratios. 
We re-implement all baselines to eliminate the discrepancy caused by MLMs and the auxiliary backward model. `$*/**$': significantly \citep{koehn2004statistical} better than the RTT with $p<0.05$ and $p<0.01$, respectively.}
\end{table*}

\subsection{Model Configuration and Pre-processing}
The MLMs and NMT models in this paper take Transformer-base \cite{vaswani2017attention} as the backbone architecture. We implement all models base on the open-source toolkit Fairseq \citep{ott2019fairseq}.  As for hyper-parameters, $\beta$ is set to 0.01 and $\gamma$ is set to 0.5 for  Zh$\rightarrow$En. For En$\rightarrow$De and En$\rightarrow$Fr, $\beta$ and $\gamma$ are set to 0.5. The replacement ratio $c$ is set to 0.2 following \citet{liu2021counterfactual}, and the candidate number $k$ is set to 1. The details of model configuration and the number of the generated adversarial examples are shown in the Appendix \ref{sec:appendixb}. 
 Following previous work, the Zh$\rightarrow$En performance is evaluated with the BLEU \citep{papineni2002bleu} score calculated by \emph{multi-bleu.perl} script. For En$\rightarrow$De and En$\rightarrow$Fr, we use SacreBLEU \citep{post2018call} for evaluation\footnote{nrefs:1 | case:mixed | eff:no | tok:intl | smooth:exp | version:2.0.0}. 

\subsection{Comparison Methods}
To test the effectiveness of our model, we take both meaning-preserving and meaning-changeable systems  as comparison methods:

\paragraph{Baseline:}
The vanilla Transformer model for NMT \citep{vaswani2017attention}. In our work, we use the baseline model to perform the forward and backward translation in the round-trip translation.

\paragraph{CharSwap:}
\citet{michel2019evaluation} craft a minor perturbation on word by swapping the internal character. They claim that character swaps have been shown to not affect human readers greatly, hence making them likely to be meaning-preserving.
\paragraph{TCWR:}
\citet{liu2021counterfactual} propose the approach of translation-counterfactual word replacement which creates augmented parallel translation corpora by random sampling new source and target phrases from the masked language models. 
\paragraph{RTT:}

\citet{zhang-etal-2021-crafting} propose to generate adversarial examples with the single round-trip translation. However, they do not provide any approach for generating the bilingual adversarial pairs. To make a fair comparison, we generate the bilingual adversarial pairs from their adversarial examples in the same way as ours.

\begin{table*}[htb]
\centering
\renewcommand\arraystretch{0.98}
\scalebox{0.95}{

\resizebox{\linewidth}{!}{
  \begin{tabular}{c|c|c c c| c|c c c |c}
        \toprule[1.2pt]
        \multirow{2}{*}{\textbf{Noise}} &\multirow{2}{*}{\textbf{Model}} &  \multicolumn{4}{c|}{\textbf{Zh$\rightarrow$En}} & \multicolumn{4}{c}{\textbf{En$\rightarrow$De}} \\
        \cmidrule(r){3-6}\cmidrule(r){7-10}
        \multicolumn{1}{c|}{}&\multicolumn{1}{c|}{} &  0.1 & 0.2 & 0.3 & \textbf{AVG} & 0.1 & 0.2 & 0.3 & \textbf{AVG}   \\
        \hline
        \multirow{5}{*}{\textbf{Deletion}} & \multicolumn{1}{l|}{baseline}  & 35.31 & 31.53 & 28.22 & 31.69 & 21.42 & 19.90 & 17.42 & 19.58  \\
        & \multicolumn{1}{l|}{+CharSwap} & 34.94 & 31.12 & 28.14 & 31.40 & 22.70 & 20.57 & 18.88 & 20.72\\
        & \multicolumn{1}{l|}{+TCWR}  & 35.02 & 31.74 & 28.45 & 31.74 &  22.45 & 20.48 & 18.66 & 20.53\\ 
        & \multicolumn{1}{l|}{+RTT} & 35.23 & 32.12 & 28.03 & 31.79 &  23.34 & 22.30	& 20.36 & 22.00  \\ 
                & \multicolumn{1}{l|}{+DRTT(ours)} &  ~~\textbf{36.63$*$} & ~~\textbf{32.96$*$}  & ~~~~\textbf{29.94$**$} & \textbf{33.18} & ~~~~\textbf{24.06}$**$ & ~~~~\textbf{23.02}$**$ & ~~~~\textbf{21.18}$**$ & \textbf{22.75} \\ 
        \hline
        \multirow{5}{*}{\textbf{Swap}} & \multicolumn{1}{l|}{baseline}  & 28.63 & 22.82 & 18.21 & 23.22 & 19.01 & 15.92 & 14.25 & 16.39  \\
        & \multicolumn{1}{l|}{+CharSwap} & 29.55 & 24.46 & 20.97 & 24.99 & 19.80 & 16.51 & 14.54 & 16.95\\
        & \multicolumn{1}{l|}{+TCWR} & 31.01 & 26.03 & 22.25 & 26.43 &  19.56 & 16.65 & 14.95 &17.05  \\ 
        & \multicolumn{1}{l|}{+RTT}  & 31.07 & 26.06 & 22.08 & 26.40 & 20.51 & 17.63 & 16.17 & 18.10  \\ 
        & \multicolumn{1}{l|}{+DRTT(ours)}  & ~~\textbf{32.03$*$} & ~~~~\textbf{26.95}$**$ & ~~~~\textbf{23.71$**$} & \textbf{27.56} &  ~~~~\textbf{21.40}$**$ & ~~~~\textbf{18.68}$**$ & ~~~~\textbf{17.53}$**$ & \textbf{19.20} \\ 
        \hline
        \multirow{5}{*}{\textbf{Insertion}} & \multicolumn{1}{l|}{baseline}  & 30.13 & 23.57 & 17.95 & 23.88 & 19.57 & 16.24 & 13.12 & 16.31 \\
        & \multicolumn{1}{l|}{+CharSwap} & 29.03 & 22.17 & 17.01  & 22.73 & 20.47 & 16.86 & 13.71 & 17.01\\
        & \multicolumn{1}{l|}{+TCWR}  & 30.12 & 23.76 & 18.02 & 23.97 &  20.73 & 17.27& \textbf{14.12} &17.37 \\ 
        & \multicolumn{1}{l|}{+RTT}  & 29.72 & 22.75 & 17.87 & 23.45 &  20.79 & 16.81	& 13.80 & 17.13\\ 
        & \multicolumn{1}{l|}{+DRTT(ours)}  & ~~~~\textbf{31.84$**$} &~~~~\textbf{24.42$**$} & ~~~~\textbf{19.43$**$} & \textbf{25.23} & ~~~~\textbf{21.24}$**$ & ~~~~\textbf{17.53}$**$ & ~~\textbf{14.12}$*$ & \textbf{17.63}  \\ 
        \hline
        \multirow{5}{*}{\textbf{Rep src}} & \multicolumn{1}{l|}{baseline}  & 33.02 & 28.15 & 23.26 &  28.14 &  20.56 & 18.40  & 16.53 & 18.50\\
        & \multicolumn{1}{l|}{+CharSwap} & 31.71 & 26.97 & 21.92 & 26.87 & 21.56 & 18.81 & 17.11 & 19.16\\
        & \multicolumn{1}{l|}{+TCWR}  & 32.83 & 28.11 & 23.38 & 28.11 & 21.43 & 19.22 & 17.10 & 19.25  \\ 
        & \multicolumn{1}{l|}{+RTT}  & 32.65 & 27.23 & 23.05 & 27.65 &  22.25 & 20.14	& 18.45 & 20.28  \\ 
        & \multicolumn{1}{l|}{+DRTT(ours)}  & ~~~~\textbf{34.76$**$} & ~~~~\textbf{29.04$**$} & ~~~~\textbf{25.06$**$} & \textbf{29.62} & ~~\textbf{22.74}$*$	& ~~\textbf{20.59}$*$ & ~~\textbf{18.87}$*$ & \textbf{20.73} \\ 
                \hline
        \multirow{5}{*}{\textbf{Rep both}} & \multicolumn{1}{l|}{baseline}  & 38.25 & 36.17 & 35.48 & 36.63 & 23.62 & 23.23 & 22.13 & 22.99 \\
        & \multicolumn{1}{l|}{+CharSwap} & 36.23 & 34.90 & 33.81 & 34.98 & 25.23 & 24.37 & 23.33 & 24.31\\
        & \multicolumn{1}{l|}{+TCWR} & 38.38 & 36.92 & 35.44 & 36.91 & 24.84 & 24.77 & 23.34 & 24.32\\ 
        & \multicolumn{1}{l|}{+RTT}  &39.13 & 36.92 & 35.23 & 37.09
        & 25.51 & 24.77 & 24.12 & 24.80  \\ 
        & \multicolumn{1}{l|}{+DRTT(ours)}  & ~~\textbf{40.07$*$} & ~~~~\textbf{38.34$**$} & ~~~~\textbf{37.22$**$} & \textbf{38.54} &  ~~~~\textbf{26.28}$**$& ~~\textbf{25.26}$*$ & ~~~~\textbf{24.87}$**$ & \textbf{25.47} \\ 

    \bottomrule
    \end{tabular}
}      
}
\caption{\label{tab:3} The RTT BLEU scores (\%) for round-trip translation on noisy test sets. `$*/**$': significantly better than RTT with $p<0.05$ and $p<0.01$,  respectively.}
\end{table*}

\section{Results and Analysis}
\subsection{Main Results}

\paragraph{Artificial Noise.}

To test robustness on noisy inputs, we follow \citet{cheng2018towards} to construct five types of synthetic perturbations with different noise ratios on the standard test set\footnote{
For each test set, we report three results with noise ratio as 0.1, 0.2 and 0.3, respectively.
Noise ratio 0.1 means 10 percent of the words in the source sentence are perturbed.}: 1) \emph{Deletion:} some words in the source sentence are randomly deleted; 2) \emph{Swap:} some words in the source sentence are randomly swapped with their right neighbors; 3) \emph{Insertion}: some words in the source sentence are randomly repeated; 4) \emph{Rep src:} short for `replacement on src'. Some words in the source sentence are  replaced with their relevant word according to the similarity of word embeddings\footnote{\url{https://github.com/Embedding/Chinese-Word-Vectors} \\ \url{https://nlp.stanford.edu/projects/glove/}
}; 5) \emph{Rep both:} short for `replacement on both'. Some words in the source sentence and their aligned target words are replaced by masked language models \footnote{Each sentence has four references on NIST test sets, we only choose sb0 for replacement.}.

Table \ref{tab:2} shows the BLEU scores of forward translation results on Zh$\rightarrow$En and En$\rightarrow$De noisy test sets. For Zh$\rightarrow$En, our approach achieves the best performance on 4 out of 5 types of noisy test sets. Compared to RTT, DRTT achieves the improvement up to 1.1 BLEU points averagely on \emph{deletion}. For En$\rightarrow$De, DRTT also performs best results on all types of noise except \emph{Rep src}. We suppose the reason is \emph{Rep src} sometimes reverses the semantics of the original sentence as we claimed above. 

Since the perturbations we introduced above may change the semantics of the source sentence, it may be problematic for us to calculate the BLEU score against the original reference sentence in Table \ref{tab:2}. Therefore, following \citet{zhang-etal-2021-crafting}, we also report the BLEU score between the source sentence and its reconstructed version through the source-target-source RTT, which is named as RTT BLEU. The intuition behind it is that: a robust NMT model translates noisy inputs well and thus has minor shifting on the round-trip translation, resulting in a high BLEU between inputs and their round-trip translation results. Following \citet{zhang-etal-2021-crafting}, we fine-tune the backward model (vanilla Transformer model) with its test set to minimize the impact of the T2S process. As shown in Table \ref{tab:3}, DRTT outperforms the meaning-preserving method and other methods on all types of noise on Zh$\rightarrow$En and En$\rightarrow$De tasks. Considering the results of Table \ref{tab:2} and Table \ref{tab:3} together, DRTT significantly improves the robustness of NMT models under various artificial noises. 

\begin{table}
\centering
 \renewcommand\arraystretch{0.98}
\scalebox{0.95}{
\begin{tabular}{l|c|c}
    \toprule
    \textbf{Method} & \textbf{En$\rightarrow$Fr} & \textbf{BLEU}$\bm \Delta$\\ 
    \midrule
    baseline & 35.02 \\
  +CharSwap & 35.59 & +0.57\\
  +TCWR & 35.64 & +0.62\\
  +RTT & 35.73 & +0.71\\
  \midrule
  +DRTT(ours) &  ~~\textbf{36.36$*$} & \textbf{+1.34}\\
    \bottomrule
    \end{tabular}
    }
\caption{\label{tab:4} The BLEU scores (\%) on the WMT'19 En$\rightarrow$Fr robustness task. `BLEU$\Delta$' denotes the gain of BLEU compared to baseline. `$*/**$': significantly better than RTT with $p<0.05$ and $p<0.01$,  respectively. }
\end{table}%

 \begin{table*}
\centering
\renewcommand\arraystretch{1}
\resizebox{1\linewidth}{!}{
  \begin{tabular}{l|c| c c c c c |c|c c}
        \toprule[1.2pt]
         \multirow{2}{*}{\textbf{Model}} &  \multicolumn{7}{c|}{\textbf{Zh$\rightarrow$En}} & \multicolumn{2}{c}{\textbf{En$\rightarrow$De}} \\
        \cmidrule(r){2-8}\cmidrule(r){9-10}
        \multicolumn{1}{c|}{}&  MT06 & MT02 & MT03 & MT04 & MT05 & MT08 & \textbf{AVG} & newstest16 & newstest17 \\
        \midrule
  baseline & 44.59 & 44.38 & 43.65 & 45.37 & 44.42 &  35.80 & 42.72 & 29.11 & 27.94\\
  +CharSwap & 43.28 & 44.80 & 44.24 & 45.52 & 43.82 & 34.29 & 42.53 & 28.48 & 27.54\\
  +TCWR & 44.55 & \textbf{45.99} &44.68 & 45.77 & 44.16 &  34.98 & 43.12 & 29.13 & 27.98\\
  +RTT &  44.62 &45.13 & 44.01 & 46.00 & \textbf{44.96}  & 35.18 & 43.06 & 29.06 & 27.42\\
  \midrule
  +DRTT(ours) & \textbf{44.76} &  45.01 & ~~~~\textbf{45.16$**$} & ~~~~\textbf{46.63$**$} & 44.78 & ~~\textbf{35.82$*$} & \textbf{43.48} & \textbf{29.30} & ~~~~\textbf{28.37$**$}\\
    \bottomrule
    \end{tabular}
}      
\caption{\label{tab:5} The BLEU scores (\%) on NIST Zh$\rightarrow$En and WMT17 En$\rightarrow$De. `$*/**$': significantly better than RTT with $p<0.05$ and $p<0.01$, respectively.}
\end{table*}

\paragraph{Natural Noise.}

In addition to the artificial noise, we also test the performance of our model on WMT'19 En$\rightarrow$Fr robustness test set which contains various noise in real-world text, e.g., exhibits typos, grammar errors, code-switching, etc.  As shown in Table \ref{tab:4}, DRTT yields improvements of 1.34 BLEU compared to the baseline, it proves that our approach also performs well in real noise scenario. Besides, DRTT achieves 0.63 BLEU improvement over RTT by filtering out 10\% of fake adversarial examples (according to Table \ref{tab:6}), which demonstrates that filtering out fake adversarial examples further improves the robustness of the model.

\subsection{Effectiveness of Adversarial Examples}
\begin{figure}
 \centering
\includegraphics[width=0.5\textwidth]{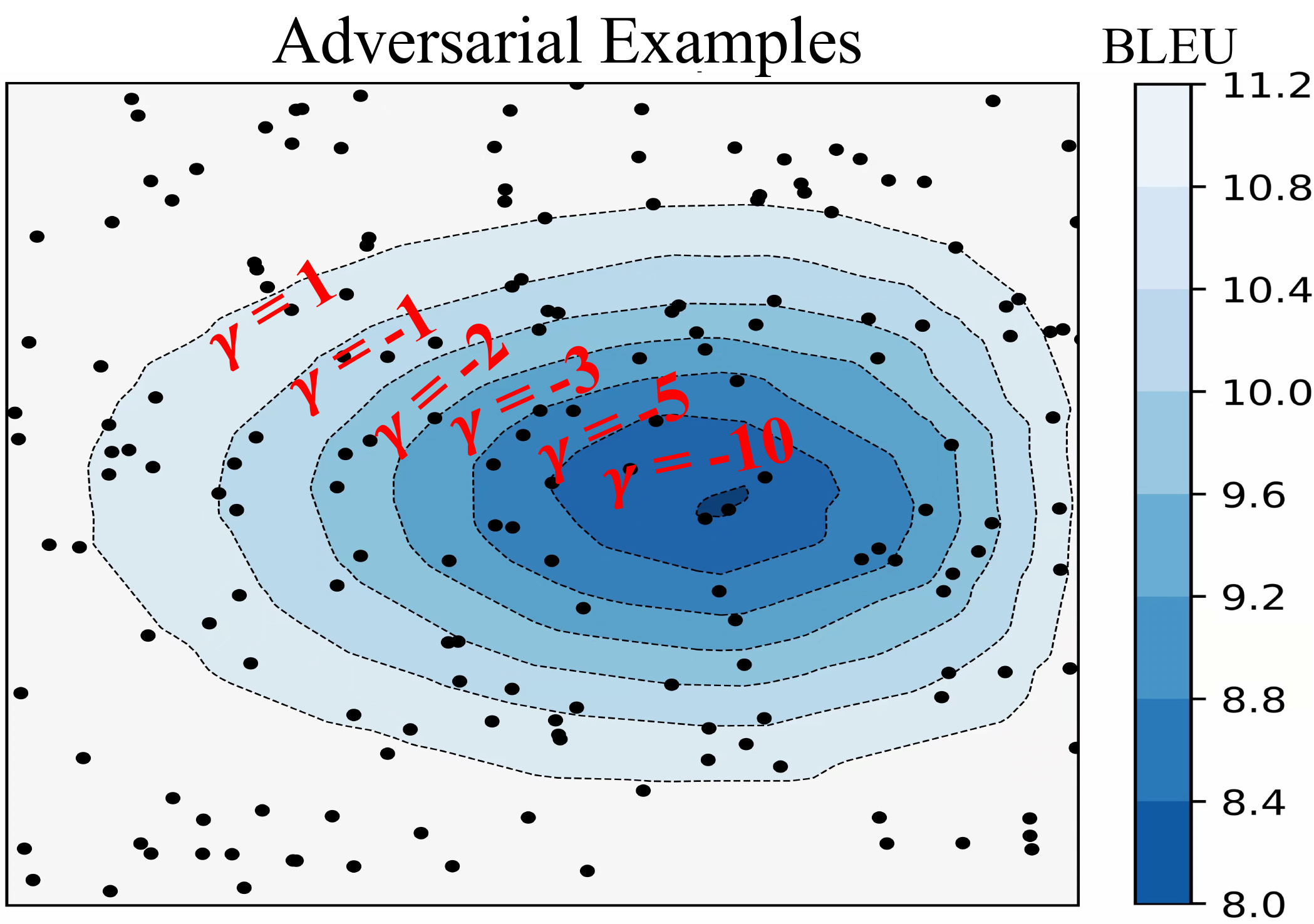}
\caption{\label{fig:3} Black spots represent the distribution of adversarial samples. The darker color indicates more effective adversarial examples generated with lower $\gamma$. } 
\end{figure}

In this sub-section, we evaluate the effectiveness of the generated adversarial examples on attacking the victim NMT model (i.e., the target NMT model without being trained on the generated adversarial pairs). In our approach, $\gamma$ in Eq.(\ref{func:4}) is a hyper-parameter
to control the strictness of our criterion on generating adversarial examples. 
Thus, we evaluate the effectiveness of adversarial examples by studying the translation performance of the victim NMT model on the set of adversarial pairs generated with different $\gamma$.
That is to say, if a sample is an adversary, it should destroy the translation performance drastically, resulting in a low BLEU score between the translation result and its paired target sentence.
The average BLEU scores of the victim model on the different adversarial pair sets (generated with $\gamma$ from -10 to 1 on NIST 06) are shown in Figure \ref{fig:3}. 
Specifically, the average BLEU on the adversarial sets generated with $\gamma=-10$ is 8.0. When we remove the restriction of $\gamma$, i.e., the DRTT is degenerated into RTT, the average BLEU for the constructed adversarial examples reaches up to 11.2.
This shows that the adversarial examples generated with lower $\gamma$ (more strict restriction) attack the model more successfully. Therefore, we can select more effective adversarial examples compared to \citet{zhang-etal-2021-crafting} by lowering the threshold $\gamma$ to create a more strict criterion.

\subsection{Clean Test set}
 
 Adding a large amount of noisy parallel data to clean training data may harm the NMT model performance on the clean test sets seriously \citep{khayrallah2018impact}. In this sub-section, we test the performance of the proposed model on the clean test sets and the results are presented in Table \ref{tab:5}. The meaning-preserving method CharSwap has negative effect on clean test set while
 DRTT achieves the best translation performance on Zh$\rightarrow$En and En$\rightarrow$De clean test sets. 
 It demonstrates that our approach not only improves the robustness of the NMT model, but also maintains its good performance on clean test sets.
 
\section{Case Study and Limitations}
\label{sec:appendixc}

In Table \ref{table:case study}, we present some cases from Zh-En adversarial pairs generated by our approach. From the case 1,  we can see  ``\begin{CJK*}{UTF8}{gbsn}{拥护}\end{CJK*}'' in the source sentence is replaced by its antonym ``\begin{CJK*}{UTF8}{gbsn}{反对}\end{CJK*}'', which reverse the meaning of the original sentence, and DRTT makes a corresponding change in the target sentence by replacing ``support'' with ``oppose''. In the other case, DRTT replaces ``\begin{CJK*}{UTF8}{gbsn}{良好}\end{CJK*}'' by its synonym ``\begin{CJK*}{UTF8}{gbsn}{不错}\end{CJK*}'', thus, ``satisfactory'' in the target sentence remains unchanged. From these cases, we find that DRTT can reasonably substitute phrases in source sequences based on the contexts and correctly modify the corresponding target phrases synchronously. 

Although the proposed approach achieves promising results, it still has limitations. A small number of authentic adversarial examples may be filtered out when the large ${\rm {d_{tgt}}}(\mathbf{y},\mathbf{y'_\delta})$ is caused by $f(\hat{x}_\delta)$, we will ameliorate this problem in the further.

 \begin{table}
 \renewcommand\arraystretch{1}
\resizebox{\linewidth}{!}{
\begin{tabular}{l}
    \toprule
x : \begin{CJK*}{UTF8}{gbsn}{\small 我们坚决拥护政府处理这一事件所采取的措施。}\end{CJK*}\\
y : we resolutely support measures taken by our \\ government in handling this incident.\\
\hline
${\rm x_\delta}$  : \begin{CJK*}{UTF8}{gbsn}{\small 我们坚决\textcolor{red}{反对}政府处理这一\textcolor{red}{案件}所采取的\textcolor{red}{举措}。}\end{CJK*}\\
 ${\rm y_\delta}$ : we resolutely \textcolor{blue}{oppose} measures taken by our \\ government in handling this \textcolor{blue}{case}. \\
    \toprule
 x : \begin{CJK*}{UTF8}{gbsn}{\small 中美双方认为 , 当前世界经济形势是良好的。通货膨胀}\end{CJK*} \\
  \begin{CJK*}{UTF8}{gbsn}{\small继续保持低水平, 大多数新兴 市场经济体的经济增长强劲。}\end{CJK*}\\
 y : china and the united states agreed that the present \\
 economic situation in the world is satisfactory, with \\
 inflation kept at a low level and most of the new market \\
 economies growing strong.\\
 \hline
 ${\rm x_\delta}$  : \begin{CJK*}{UTF8}{gbsn}{\small \textcolor{red}{俄}美双方认为, 当前世界\textcolor{red}{贸易势头}是\textcolor{red}{不错}的。 通货膨胀}\end{CJK*}\\
 \begin{CJK*}{UTF8}{gbsn}{\small 继续保持低\textcolor{red}{速度}, 大多数新兴市场经济体的经济\textcolor{red}{发展}强劲。}\end{CJK*}\\
 ${\rm y_\delta}$ :  \textcolor{blue}{russia} and the united states agreed that the present \\
 \textcolor{blue}{trade trend} in the world is satisfactory, with inflation \\
 kept at a low \textcolor{blue}{rate} and most of the new market economies \\
 \textcolor{blue}{developing} strong. \\
    \toprule
\end{tabular}
}
\caption{Case study for the proposed approach. The words in red and blue color represents the augmented words on the source and target side, respectively. }

\label{table:case study} 
\end{table}

\section{Conclusion and Future Work}
We propose a new criterion for NMT adversarial examples based on Doubly Round-Trip Translation, which can ensure the examples that meet our criterion are the authentic adversarial examples. Additionally, based on this criterion, we introduce the masked language models to generate bilingual adversarial pairs, which can be used to improve the robustness of the NMT model substantially. Extensive experiments on both the clean and noisy test sets show that our approach not only improves the robustness of the NMT model but also performs well on the clean test sets. In future work, we will refine the limitations of this work and then explore to improve the robustness of forward and backward models simultaneously.
We hope our work will provide a new perspective for future researches on adversarial examples.

\section*{Acknowledgements}
The research work descried in this paper has been supported by the National Key R\&D Program of China (2020AAA0108001) and the National Nature Science Foundation of China (No. 61976016, 61976015, and 61876198). The authors would like to thank the anonymous reviewers for their valuable comments and suggestions to improve this paper.

\bibliography{anthology,custom}

\begin{thebibliography}{28}
\expandafter\ifx\csname natexlab\endcsname\relax\def\natexlab#1{#1}\fi

\bibitem[{Bahdanau et~al.(2014)Bahdanau, Cho, and Bengio}]{bahdanau2014neural}
Dzmitry Bahdanau, Kyunghyun Cho, and Yoshua Bengio. 2014.
\newblock Neural machine translation by jointly learning to align and
  translate.
\newblock \emph{arXiv preprint arXiv:1409.0473}.

\bibitem[{Belinkov and Bisk(2017)}]{belinkov2017synthetic}
Yonatan Belinkov and Yonatan Bisk. 2017.
\newblock Synthetic and natural noise both break neural machine translation.
\newblock \emph{arXiv preprint arXiv:1711.02173}.

\bibitem[{Chen et~al.(2021)Chen, Fan, Zhang, Chen, and
  Huang}]{chen2021manifold}
Guandan Chen, Kai Fan, Kaibo Zhang, Boxing Chen, and Zhongqiang Huang. 2021.
\newblock Manifold adversarial augmentation for neural machine translation.
\newblock In \emph{Findings of the Association for Computational Linguistics:
  ACL-IJCNLP 2021}, pages 3184--3189.

\bibitem[{Cheng et~al.(2019)Cheng, Jiang, and Macherey}]{cheng2019robust}
Yong Cheng, Lu~Jiang, and Wolfgang Macherey. 2019.
\newblock Robust neural machine translation with doubly adversarial inputs.
\newblock In \emph{Proceedings of the 57th Annual Meeting of the Association
  for Computational Linguistics}, pages 4324--4333.

\bibitem[{Cheng et~al.(2020)Cheng, Jiang, Macherey, and
  Eisenstein}]{cheng-etal-2020-advaug}
Yong Cheng, Lu~Jiang, Wolfgang Macherey, and Jacob Eisenstein. 2020.
\newblock \href {https://doi.org/10.18653/v1/2020.acl-main.529} {{A}dv{A}ug:
  Robust adversarial augmentation for neural machine translation}.
\newblock In \emph{Proceedings of the 58th Annual Meeting of the Association
  for Computational Linguistics}, pages 5961--5970, Online. Association for
  Computational Linguistics.

\bibitem[{Cheng et~al.(2018)Cheng, Tu, Meng, Zhai, and Liu}]{cheng2018towards}
Yong Cheng, Zhaopeng Tu, Fandong Meng, Junjie Zhai, and Yang Liu. 2018.
\newblock Towards robust neural machine translation.
\newblock \emph{arXiv preprint arXiv:1805.06130}.

\bibitem[{Cho et~al.(2014)Cho, van Merrienboer, G{\"u}l{\c{c}}ehre, Bahdanau,
  Bougares, Schwenk, and Bengio}]{cho2014learning}
Kyunghyun Cho, Bart van Merrienboer, {\c{C}}aglar G{\"u}l{\c{c}}ehre, Dzmitry
  Bahdanau, Fethi Bougares, Holger Schwenk, and Yoshua Bengio. 2014.
\newblock Learning phrase representations using rnn encoder-decoder for
  statistical machine translation.
\newblock In \emph{EMNLP}.

\bibitem[{Conneau and Lample(2019)}]{conneau2019cross}
Alexis Conneau and Guillaume Lample. 2019.
\newblock Cross-lingual language model pretraining.
\newblock \emph{Advances in Neural Information Processing Systems},
  32:7059--7069.

\bibitem[{Devlin et~al.(2018)Devlin, Chang, Lee, and Toutanova}]{bert}
Jacob Devlin, Ming{-}Wei Chang, Kenton Lee, and Kristina Toutanova. 2018.
\newblock \href {http://arxiv.org/abs/1810.04805} {{BERT:} pre-training of deep
  bidirectional transformers for language understanding}.
\newblock \emph{CoRR}, abs/1810.04805.

\bibitem[{Dyer et~al.(2013)Dyer, Chahuneau, and Smith}]{dyer2013simple}
Chris Dyer, Victor Chahuneau, and Noah~A Smith. 2013.
\newblock A simple, fast, and effective reparameterization of ibm model 2.
\newblock In \emph{Proceedings of the 2013 Conference of the North American
  Chapter of the Association for Computational Linguistics: Human Language
  Technologies}, pages 644--648.

\bibitem[{Ebrahimi et~al.(2018)Ebrahimi, Rao, Lowd, and
  Dou}]{ebrahimi2018hotflip}
Javid Ebrahimi, Anyi Rao, Daniel Lowd, and Dejing Dou. 2018.
\newblock Hotflip: White-box adversarial examples for text classification.
\newblock In \emph{Proceedings of the 56th Annual Meeting of the Association
  for Computational Linguistics (Volume 2: Short Papers)}, pages 31--36.

\bibitem[{Jiao et~al.(2019)Jiao, Yin, Shang, Jiang, Chen, Li, Wang, and
  Liu}]{jiao2019tinybert}
Xiaoqi Jiao, Yichun Yin, Lifeng Shang, Xin Jiang, Xiao Chen, Linlin Li, Fang
  Wang, and Qun Liu. 2019.
\newblock Tinybert: Distilling bert for natural language understanding.
\newblock \emph{arXiv preprint arXiv:1909.10351}.

\bibitem[{Karpukhin et~al.(2019)Karpukhin, Levy, Eisenstein, and
  Ghazvininejad}]{karpukhin2019training}
Vladimir Karpukhin, Omer Levy, Jacob Eisenstein, and Marjan Ghazvininejad.
  2019.
\newblock Training on synthetic noise improves robustness to natural noise in
  machine translation.
\newblock \emph{arXiv preprint arXiv:1902.01509}.

\bibitem[{Khayrallah and Koehn(2018)}]{khayrallah2018impact}
Huda Khayrallah and Philipp Koehn. 2018.
\newblock On the impact of various types of noise on neural machine
  translation.
\newblock In \emph{Proceedings of the 2nd Workshop on Neural Machine
  Translation and Generation}, pages 74--83.

\bibitem[{Kingma and Ba(2014)}]{kingma2014adam}
Diederik~P Kingma and Jimmy Ba. 2014.
\newblock Adam: A method for stochastic optimization.
\newblock \emph{arXiv preprint arXiv:1412.6980}.

\bibitem[{Koehn(2004)}]{koehn2004statistical}
Philipp Koehn. 2004.
\newblock Statistical significance tests for machine translation evaluation.
\newblock In \emph{Proceedings of the 2004 conference on empirical methods in
  natural language processing}, pages 388--395.

\bibitem[{Liu et~al.(2021)Liu, Kusner, and Blunsom}]{liu2021counterfactual}
Qi~Liu, Matt Kusner, and Phil Blunsom. 2021.
\newblock Counterfactual data augmentation for neural machine translation.
\newblock In \emph{Proceedings of the 2021 Conference of the North American
  Chapter of the Association for Computational Linguistics: Human Language
  Technologies}, pages 187--197.

\bibitem[{Lu et~al.(2021)Lu, Zeng, Zhang, Wu, and Li}]{lu-etal-2021-attention}
Yu~Lu, Jiali Zeng, Jiajun Zhang, Shuangzhi Wu, and Mu~Li. 2021.
\newblock \href {https://doi.org/10.18653/v1/2021.acl-long.103} {Attention
  calibration for transformer in neural machine translation}.
\newblock In \emph{Proceedings of the 59th Annual Meeting of the Association
  for Computational Linguistics and the 11th International Joint Conference on
  Natural Language Processing (Volume 1: Long Papers)}, pages 1288--1298,
  Online. Association for Computational Linguistics.

\bibitem[{Michel et~al.(2019)Michel, Li, Neubig, and
  Pino}]{michel2019evaluation}
Paul Michel, Xian Li, Graham Neubig, and Juan Pino. 2019.
\newblock On evaluation of adversarial perturbations for sequence-to-sequence
  models.
\newblock In \emph{Proceedings of the 2019 Conference of the North American
  Chapter of the Association for Computational Linguistics: Human Language
  Technologies, Volume 1 (Long and Short Papers)}, pages 3103--3114.

\bibitem[{Michel and Neubig(2018)}]{michel2018mtnt}
Paul Michel and Graham Neubig. 2018.
\newblock Mtnt: A testbed for machine translation of noisy text.
\newblock In \emph{Proceedings of the 2018 Conference on Empirical Methods in
  Natural Language Processing}, pages 543--553.

\bibitem[{Niu et~al.(2020)Niu, Mathur, f~Dinu, and
  Al-Onaizan}]{niu-etal-2020-evaluating}
Xing Niu, Prashant Mathur, Georgiana f~Dinu, and Yaser Al-Onaizan. 2020.
\newblock \href {https://doi.org/10.18653/v1/2020.acl-main.755} {Evaluating
  robustness to input perturbations for neural machine translation}.
\newblock In \emph{Proceedings of the 58th Annual Meeting of the Association
  for Computational Linguistics}, pages 8538--8544, Online. Association for
  Computational Linguistics.

\bibitem[{Ott et~al.(2019)Ott, Edunov, Baevski, Fan, Gross, Ng, Grangier, and
  Auli}]{ott2019fairseq}
Myle Ott, Sergey Edunov, Alexei Baevski, Angela Fan, Sam Gross, Nathan Ng,
  David Grangier, and Michael Auli. 2019.
\newblock fairseq: A fast, extensible toolkit for sequence modeling.
\newblock In \emph{Proceedings of NAACL-HLT 2019: Demonstrations}.

\bibitem[{Papineni et~al.(2002)Papineni, Roukos, Ward, and
  Zhu}]{papineni2002bleu}
Kishore Papineni, Salim Roukos, Todd Ward, and Wei-Jing Zhu. 2002.
\newblock Bleu: a method for automatic evaluation of machine translation.
\newblock In \emph{Proceedings of the 40th annual meeting of the Association
  for Computational Linguistics}, pages 311--318.

\bibitem[{Post(2018)}]{post2018call}
Matt Post. 2018.
\newblock A call for clarity in reporting bleu scores.
\newblock \emph{arXiv preprint arXiv:1804.08771}.

\bibitem[{Sennrich et~al.(2016)Sennrich, Haddow, and
  Birch}]{sennrich2016neural}
Rico Sennrich, Barry Haddow, and Alexandra Birch. 2016.
\newblock Neural machine translation of rare words with subword units.
\newblock In \emph{Proceedings of the 54th Annual Meeting of the Association
  for Computational Linguistics (Volume 1: Long Papers)}, pages 1715--1725.

\bibitem[{Vaswani et~al.(2017)Vaswani, Shazeer, Parmar, Uszkoreit, Jones,
  Gomez, Kaiser, and Polosukhin}]{vaswani2017attention}
Ashish Vaswani, Noam Shazeer, Niki Parmar, Jakob Uszkoreit, Llion Jones,
  Aidan~N Gomez, {\L}ukasz Kaiser, and Illia Polosukhin. 2017.
\newblock Attention is all you need.
\newblock In \emph{Advances in neural information processing systems}, pages
  5998--6008.

\bibitem[{Yang et~al.(2020)Yang, Hu, Han, Huang, and Ju}]{yang2020csp}
Zhen Yang, Bojie Hu, Ambyera Han, Shen Huang, and Qi~Ju. 2020.
\newblock Csp: Code-switching pre-training for neural machine translation.
\newblock In \emph{Proceedings of the 2020 Conference on Empirical Methods in
  Natural Language Processing (EMNLP)}, pages 2624--2636.

\bibitem[{Zhang et~al.(2021)Zhang, Zhang, Chen, and
  He}]{zhang-etal-2021-crafting}
Xinze Zhang, Junzhe Zhang, Zhenhua Chen, and Kun He. 2021.
\newblock \href {https://doi.org/10.18653/v1/2021.acl-long.153} {Crafting
  adversarial examples for neural machine translation}.
\newblock In \emph{Proceedings of the 59th Annual Meeting of the Association
  for Computational Linguistics and the 11th International Joint Conference on
  Natural Language Processing (Volume 1: Long Papers)}, pages 1967--1977,
  Online. Association for Computational Linguistics.

\end{thebibliography}
\bibliographystyle{acl_natbib}

\appendix
\clearpage
\section{Bilingual Adversarial Pair Generation}
\label{sec:appendixa}
\begin{algorithm}

\caption{Bilingual Adversarial Pair Generation} \label{algorithm:alg1}
\KwIn{ A sequence pair $(x,y)$, a sampling probability $c$, an alignment mapping $p$, candidate words $k$, masked language models M-MLM and T-MLM, thresholds $\beta$ and $\gamma$. }
\KwOut{A bilingual adversarial pair $(x_\delta,y_\delta)$}
\SetKwFunction{FMain}{BilAdvGen}
\SetKwProg{Fn}{Function}{:}{}
\Fn{\FMain{$x,y$}}{
\While{$n \leq len(x)*c$ }{
{$r_i^j \leftarrow $ M-MLM $(x^{\backslash i})$;}\\
$x^{\backslash i:j}\leftarrow$ Replace$(x, r_i^j)$ \\
$r_i^* \leftarrow \arg\max$  $ {\rm d_{src}} (x, x^{\backslash i:j})$ (\ref{func:2});\\
$x^{\backslash i:*} \leftarrow$ Replace$(x, r_i^*)$ \\
Get aligned index $p(i)$;\\
$w_{p(i)} \leftarrow$ T-MLM $(x^{\backslash i:*},y^{\backslash p(i)})$;\\
$y_\delta \leftarrow$ Replace$(y, w_{p(i)})$ \\
$n\leftarrow n+1$ }
}
\If {${\rm d_{src}}(x,x_\delta)> \beta$ {\rm and} ${\rm d_{tgt}}(y,y'_\delta)< \gamma$}{
\textbf{return} $x_\delta,y_\delta$}
\end{algorithm}

\section{Implementation Details}
\label{sec:appendixb}
 As for Zh$\rightarrow$En, we apply the separate byte-pair encoding (BPE) \citep{sennrich2016neural} encoding with 30K merge operations for Zh and En, respectively, the peak learning rate of 5e-4, and the training step is 100K. For En$\rightarrow$De and En$\rightarrow$Fr, we apply the joint BPE with 32K merge operations, the learning rate of 7e-4 and the training step is 200K. The dropout ratio is 0.1. We use Adam optimizer \cite{kingma2014adam} with 4k warm-up steps. All models are trained on 8 NVIDIA Tesla V100 (32GB) GPUs.
\begin{table}[h]
\centering
 \renewcommand\arraystretch{1}
\scalebox{1}{
\begin{tabular}{l|c|c|c}
    \toprule
    \textbf{Method} & \textbf{Zh$\rightarrow$En} & \textbf{En$\rightarrow$De} & \textbf{En$\rightarrow$Fr}\\ 
    \midrule
    original & 1252977 & 5859951 & 2037962 \\
    -CharSwap & 1252977 & 5859951 & 2037962 \\
    -TCWR & 1252977 & 5859951 & 2037962 \\
    -RTT &1236485 & 2670044 & 1639661 \\
    -DRTT(ours) &956308 & 2336285  & 1466756\\
    \bottomrule
    \end{tabular}
    }
\caption{\label{tab:6} The statistics of the number of adversarial examples generated by different methods. }
\end{table}%

\end{document}